\documentclass{article} % For LaTeX2e
\usepackage{iclr2024_conference,times}
\usepackage{amsmath,amsfonts,bm}

\def\eqref#1{equation~\ref{#1}}
\def\1{\bm{1}}

\DeclareMathAlphabet{\mathsfit}{\encodingdefault}{\sfdefault}{m}{sl}
\SetMathAlphabet{\mathsfit}{bold}{\encodingdefault}{\sfdefault}{bx}{n}

\usepackage{hyperref}
\usepackage{url}
\usepackage{color}
\usepackage{subfigure}
\usepackage{graphicx}
\usepackage{multirow}
\usepackage{array}
\usepackage{booktabs}
\usepackage{ulem}
\usepackage{enumitem}
\usepackage{algpseudocode}
\usepackage{algorithmicx,algorithm}

\makeatletter
\newcommand{\printfnsymbol}[1]{%
  \textsuperscript{\@fnsymbol{#1}}%
}
\makeatother

\makeatletter
\def\@fnsymbol#1{\ensuremath{\ifcase#1\or \dagger\or \ddagger\or
   \mathsection\or \mathparagraph\or \|\or **\or \dagger\dagger
   \or \ddagger\ddagger \else\@ctrerr\fi}}
\makeatother

\title{Adversarial Robustness for Visual Grounding of Multimodal Large Language Models}

\author{Kuofeng Gao\textsuperscript{\rm 1},\ \ Yang Bai\textsuperscript{\rm 2},\ \ Jiawang Bai\textsuperscript{\rm 1},\ \ Yong Yang\textsuperscript{\rm 2}\thanks{Corresponding authors},\ \ Shu-Tao Xia\textsuperscript{\rm 1,3}\printfnsymbol{1}  \\
\textsuperscript{\rm 1} Tsinghua University \quad \textsuperscript{\rm 2} Tencent Security Platform \quad  \textsuperscript{\rm 3} Peng Cheng Laboratory \\
\texttt{\{gkf21,bjw19\}@mails.tsinghua.edu.cn, mavisbai@tencent.com}\\
\texttt{coolcyang@tencent.com, xiast@sz.tsinghua.edu.cn}
}

\iclrfinalcopy % Uncomment for camera-ready version, but NOT for submission.
\begin{document}

\maketitle

\begin{abstract}
Multi-modal Large Language Models (MLLMs) have recently achieved enhanced performance across various vision-language tasks including visual grounding capabilities. However, the adversarial robustness of visual grounding remains unexplored in MLLMs. To fill this gap, we use referring expression comprehension (REC) as an example task in visual grounding and propose three adversarial attack paradigms as follows. Firstly, untargeted adversarial attacks induce MLLMs to generate incorrect bounding boxes for each object. Besides, exclusive targeted adversarial attacks cause all generated outputs to the same target bounding box. In addition, permuted targeted adversarial attacks aim to permute all bounding boxes among different objects within a single image. Extensive experiments demonstrate that the proposed methods can successfully attack visual grounding capabilities of MLLMs. Our methods not only provide a new perspective for designing novel attacks but also serve as a strong baseline for improving the adversarial robustness for visual grounding of MLLMs.

% Specifically, untargeted attacks prompt MLLMs to generate incorrect bounding boxes for each object, exclusive targeted attacks cause all outputs to converge to the same bounding box, and permuted targeted attacks aim to permute all bounding boxes within a single image. Extensive experiments confirm the effectiveness of our proposed methods in attacking MLLMs' visual grounding capabilities. These methods not only offer a fresh perspective for designing novel attacks, but also establish a strong baseline for enhancing the adversarial robustness of MLLMs in visual grounding tasks.
% However, the deployment of VLMs necessitates substantial energy consumption and computational resources. Once attackers maliciously induce high energy consumption and latency time (energy-latency cost) during inference  of VLMs, it will exhaust computational resources.
% , which exposes a new attack surface for attackers. Specifically, they can maliciously induce high energy consumption and latency time (energy-latency cost) during inference stage of VLMs to exhaust computational resources.
% amplifies the risk of service being maliciously occupied. 
% To mitigate this risk, it is important to evaluate the worst-case energy consumption and latency time. 
% In this paper, we explore this attack surface about availability of VLMs and aim to induce high energy-latency cost during inference of VLMs. We find that high energy-latency cost during inference of VLMs can be manipulated by maximizing the length of generated sequences. To this end, 
\end{abstract}

% As a result, VLMs have become an essential component in various aphis substantial increase in the length of generated sequences poses a potential challenge for various applications.plications, enabling more intelligent and context-aware interactions between humans and machines

\section{Introduction}
\label{sec:intro}
Multi-modal Large Language Models (MLLMs) \citep{alayrac2022flamingo,chen2022visualgpt,liu2023visual,li2021align,li2023blip}, such as GPT-4 \citep{openai2023gpt4}, integrate visual modality into large language models (LLMs)
% , enabling them to process both visual and textual information simultaneously. This fusion of modalities allows MLLMs to achieve a deeper  comprehension of relationships between diverse data types, resulting in 
and have achieved state-of-the-art performance across various multi-modal tasks, including image captioning and visual question answering. Recent advancements in research \citep{chen2023minigpt,chen2023shikra,peng2023kosmos} have further unlocked the potential visual grounding capabilities of MLLMs. Through this grounding capability, MLLMs can accurately recognize objects, locate them, and provide visual responses, such as bounding boxes, thereby facilitating additional vision-language tasks, including referring expression comprehension.

Despite the impressive multi-modal performance of MLLMs, recent studies \citep{dong2023robust,zhao2023evaluating,carlini2023aligned,qi2023visual,gao2024inducing,gao2024energy,yang2024cheating} have revealed their susceptibility of MLLMs against adversarial attacks. Adversarial attacks manipulate input data with an imperceptible perturbation with the intention of misleading the model, often resulting in incorrect outputs. Most existing adversarial attacks on MLLMs have made main efforts on the image captioning and visual question answering task. Specifically, they craft an adversarial image that closely resembles the original image and employ it to prompt MLLMs, which can induce MLLMs to generate a wrong caption or reply an incorrect answer. However, the adversarial robustness on visual grounding is still unclear.

In this paper, we study the impact of adversarial attacks on visual grounding capabilities of MLLMs at first. As a representative example, we evaluate the adversarial robustness for visual grounding of MLLMs specifically through the task of referring expression comprehension. Referring expression comprehension (REC) is the process of identifying and localizing objects within an image based on a given textual prompt, ultimately generating bounding boxes of objects. Following previous work \citep{dong2023robust,zhao2023evaluating}, we focus on visual modality and aim to craft adversarial images with an imperceptible perturbation to perform adversarial attacks. 

Concretely, three attack paradigms are proposed tailored for REC of MLLMs as follows. Firstly, an untargeted attack aims to reduce the accuracy of bounding box predictions. This attack is deemed successful if the objects in the adversarial images are incorrectly located based on the original textual prompt. Besides, based on the type of target bounding box, two categories of targeted adversarial attacks are proposed, \textit{i.e.}, exclusive targeted adversarial attacks and permuted targeted adversarial attacks. Exclusive targeted adversarial attacks deceive MLLMs to generate the same target bounding box, such as top left corner, regardless of their ground-truths. In contrast, permuted targeted adversarial attacks assign different target bounding boxes to different objects with the attacking goal of rearranging all bounding boxes within a single image.

The main contributions of this work are three-fold: (1) To the best of our knowledge, we are the first to reveal the adversarial threat in visual grounding of MLLMs. (2) We propose three attack paradigms to evaluate grounding adversarial robustness of MLLMs, including untargeted adversarial attacks, exclusive targeted adversarial attacks and permuted targeted adversarial attacks. (3) Extensive experiments are conducted, which verify the effectiveness of our proposed attacks.

\begin{figure*}[t] \centering     
\includegraphics[width=0.98\columnwidth]{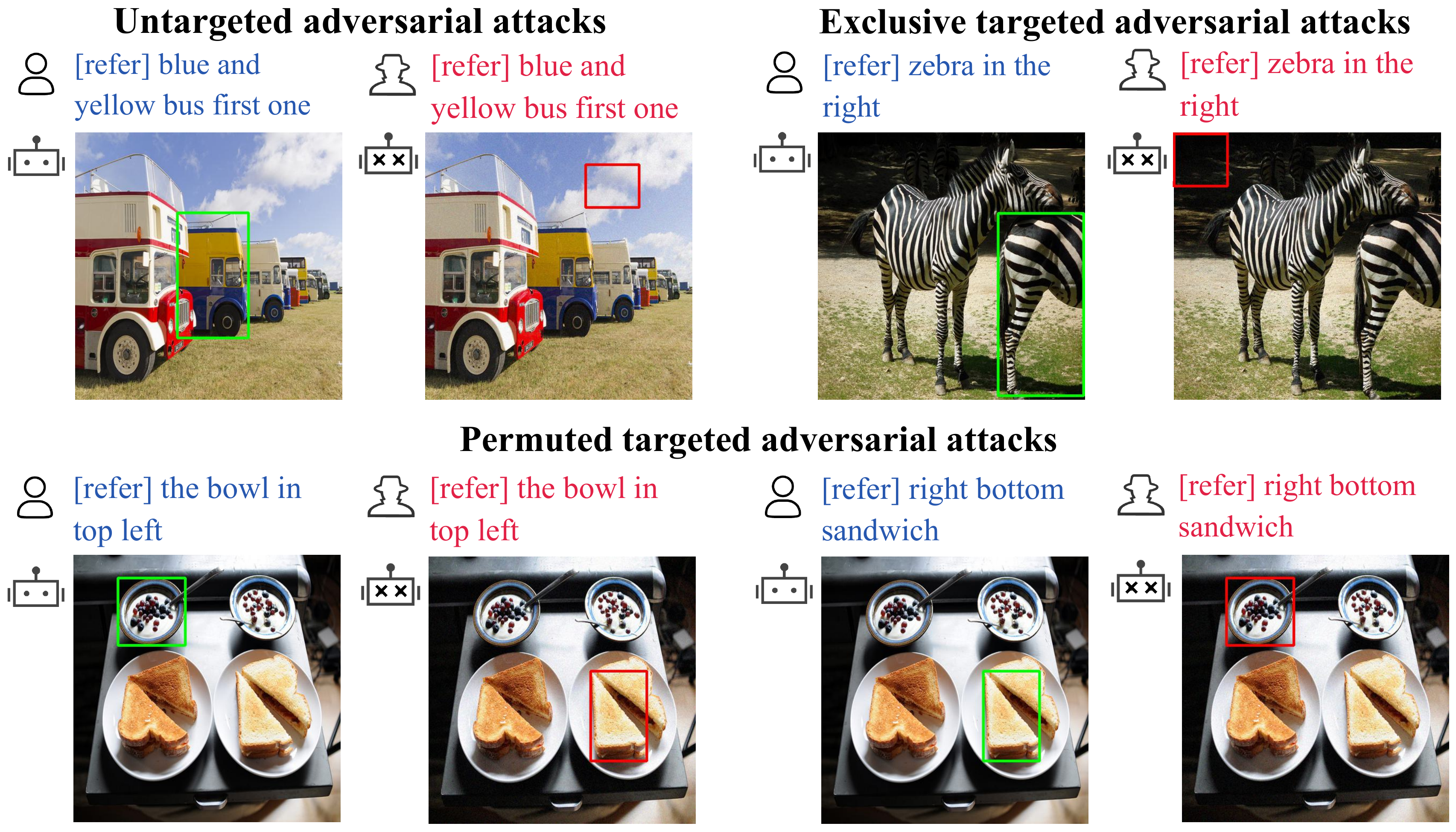} 
\vspace{-1em}
\caption{Three adversarial attack paradigms are proposed to evaluate the adversarial robustness for visual grounding of MLLMs. } 
\vspace{-1em}
\label{main}
\end{figure*}

\section{The Proposed Attack}
\label{sec:proposed attack}

\subsection{Preliminaries}
Given an image $\bm{x}$ and multiple input textual prompts $T=\{t_i\}_{i=1}^N$, referring expression comprehension (REC) aims to locate corresponding target objects by bounding boxes $B=\{b_i\}_{i=1}^N$. 
During training of MLLMs, these bounding boxes are transformed into the textual formatting and MLLMs are trained using the auto-regressive loss.
Successful REC by MLLMs is assumed when Intersection over Union (IoU) between the ground-truth and predicted bounding boxes exceeds 0.5.

% \textbf{Goals and capabilities.} 
\textbf{Threat model.} The goal of attackers is to optimize an imperceptible perturbation to craft adversarial images $\hat{\bm{x}}$ to achieve adversarial attacks. Specifically, the involved perturbation  is restricted within a predefined magnitude $\epsilon$ in $l_{\infty}$ norm, ensuring it difficult to detect.
% \textbf{Knowledge and background.} 
As suggested in~\citet{bagdasaryan2023ab,qi2023visual}, we assume that the victim MLLMs can be accessed in full knowledge, including both architectures and parameters of victim MLLMs. 

\subsection{Untargeted Adversarial Attacks}
Untargeted adversarial attacks craft adversarial images $\hat{\bm{x}}$ with the aim of causing MLLMs to predict a bounding box that deviates from its ground-truth $b_i$ when given an input textual prompt $t_i$. To this end, we propose two methods to mislead the MLLM's predictions, \textit{i.e.}, \textbf{image embedding attack} and \textbf{textual bounding box attack}. 

\textbf{Image embedding attack}. MLLMs first use vision encoders $f(\cdot)$ to extract image embeddings and generate the textual formatting of bounding boxes. Hence, image embedding attack can be implemented by maximizing the $l_2$ distance of the image embeddings between the original image $\bm{x}$ and the adversarial image $\hat{\bm{x}}$. The disrupted image embeddings will result in the model's inability to accurately predict the bounding boxes based on the input textual prompts. The objective function can be formulated as:
\begin{equation}
\begin{aligned}
\max_{\bm{x}} ||f(\hat{\bm{x}})-f(\bm{x}) ||_2^2, \quad \text{s.t.}\ ||\hat{\bm{x}}-\bm{x}||_{\infty} \leq \epsilon.
\end{aligned}
\label{eq:image embedding attack}
\end{equation}
% where $\epsilon$ denotes the maximum perturbation strength.
% , which can ensure the imperceptibility of adversarial perturbations.

\textbf{Textual bounding box attack}. Based on the original image $\bm{x}$ and the input textual prompt $t_{i}$, MLLMs $g(\cdot)$ will generate the textual formatting of the bounding box $b_{i}$ in an auto-regressive manner. Concretely, MLLMs aim to estimate the probability of a next token given its context, including the original image $\bm{x}$, the input textual prompt $t_{i}$, and previous generated $M$ tokens. Given the textual formatting of ground-truth bounding box $b_{i}=\{b_{i}^j\}_{j=1}^L$, the objective function can be formulated as:
\begin{equation}
\begin{aligned}
\min_{\bm{x}} \sum_{j=1}^L \text{log}\ p_g(b_i^j \mid b_i^{j < M};\ \bm{x};\ t_{i}), \quad \text{s.t.}\ ||\hat{\bm{x}}-\bm{x}||_{\infty} \leq \epsilon,
\end{aligned}
\label{eq:textual bounding box attack}
\end{equation}
where $b_i^{j < M}$ denotes the previous generated $M$ tokens. Textual bounding box attacks minimize the log-likelihood of the textual formatting of the ground-truth bounding box.
% to achieve the untargeted adversarial attacks.

\subsection{Targeted Adversarial Attacks}
Targeted adversarial attacks craft adversarial images $\hat{\bm{x}}$ with the goal of causing MLLMs to predict a target bounding box different from the ground-truth bounding box $b_i$ when given an input textual prompt $t_i$. Based on the type of target bounding box, two targeted attack paradigms are proposed, including \textbf{exclusive targeted adversarial attacks} and \textbf{permuted targeted adversarial attacks}. 

\textbf{Exclusive targeted adversarial attacks}. Regardless of the input textual prompt, exclusive targeted adversarial attacks deceive MLLMs to locate all objects in images to the same target bounding box, denoted as $b_u$. To achieve this attack, given the textual formatting of target bounding box $b_{u}=\{b_{u}^j\}_{j=1}^L$, the objective function can be formulated as:
\begin{equation}
\begin{aligned}
\max_{\bm{x}} \sum_{j=1}^L \text{log}\ p_g(b_u^j \mid b_u^{j < M};\ \bm{x};\ t_{i}), \quad \text{s.t.}\ ||\hat{\bm{x}}-\bm{x}||_{\infty} \leq \epsilon,
\end{aligned}
\label{eq:exclusive targeted adversarial attack}
\end{equation}
where $b_u^{j < M}$ denotes previous generated $M$ tokens. Exclusive targeted adversarial attacks maximize the log-likelihood of the textual formatting of the same target bounding box.

\textbf{Permuted targeted adversarial attacks}. Permuted targeted adversarial attacks aim to rearrange bounding box of all objects within an image. The target bounding box is determined based on the ground-truth bounding box. Given an input textual prompt $t_i$ associated with the corresponding bounding box $b_i$, permuted targeted adversarial attacks set the target bounding box as $b_{(i+1)\ \text{mod}\ N}$, where $N$ represents the number of objects within the image. This approach ensures that each object's bounding box is shifted to the next object, effectively rearranging all bounding boxes in the image. The objective function can be formulated as:
\begin{equation}
\begin{aligned}
\max_{\bm{x}} \sum_{j=1}^L \text{log}\ p_g(b_{(i+1)\ \text{mod}\ N}^j \mid b_{(i+1)\ \text{mod}\ N}^{j < M};\ \bm{x};\ t_{i}), \quad \text{s.t.}\ ||\hat{\bm{x}}-\bm{x}||_{\infty} \leq \epsilon,
\end{aligned}
\label{eq:permuted targeted adversarial attack}
\end{equation}
where $L$ denotes the token number of textual formatting of target bounding box and $b_{(i+1)\ \text{mod}\ N}^{j < M}$ denotes previous generated $M$ tokens. Permuted targeted adversarial attacks maximize the log-likelihood of the textual formatting of the target bounding box, which is shifted from another object within an image.

\section{Experiments}
\subsection{Experimental Setups}
\textbf{Models and datasets.} 
We consider the 7B version of MiniGPT-v2 \cite{chen2023minigpt} as the sandbox to launch our attack.  Moreover, RefCOCO \citep{kazemzadeh2014referitgame} and RefCOCO+ \citep{yu2016modeling}, and RefCOCOg \citep{mao2016generation} are considered as benchmark datasets for evaluation.
% To test the effectiveness of our attack methods for referring expression comprehension task, we conduct evaluations on  benchmark datasets.

% We consider four open-source and advanced large vision-language models as our evaluation benchmark, including BLIP \citep{li2022blip}, BLIP-2 \citep{li2023blip}, InstructBLIP \citep{dai2023instructblip}, and MiniGPT-4 \citep{zhu2023minigpt}.  Concretely, we adopt the BLIP with the basic multi-modal mixture of encoder-decoder model in 224M version, BLIP-2 with an OPT-2.7B LM \citep{zhang2022opt}, InstructBLIP and MiniGPT-4 with a Vicuna-7B LM \citep{chiang2023vicuna}. These models perform the captioning task for the image under their default prompt template. Results of more tasks are in Appendix \ref{sec: More tasks}. We randomly choose the 1,000 images from MS-COCO \citep{lin2014microsoft} and ImageNet \citep{deng2009imagenet} dataset, respectively, as our evaluation dataset. More details about target models are shown in Appendix \ref{sec: Target models}.

\textbf{Baselines and setups.}
To optimize three proposed adversarial attacks, we perform the projected gradient descent (PGD) \citep{madry2017towards} algorithm in $T=100$ iterations. Besides, the perturbation magnitude is set as $\epsilon=16$ within $l_{\infty}$ restriction, following \citet{dong2023robust,qi2023visual}, and the step size is set as $\alpha=1$. In exclusive targeted adversarial attacks, the top left corner, which accounts for 4\% of the total area is set as the target bounding box.

\textbf{Evaluation metrics.} We employ Intersection over Union (IoU) with a threshold of 0.5 (IoU@0.5) as the evaluation metric. A prediction is considered correct if the IoU between the predicted and ground-truth bounding boxes is greater than 0.5. 
For untargeted adversarial attacks, a lower IoU@0.5 value indicates a more effective attack. Conversely, for the two proposed targeted adversarial attacks, a higher IoU@0.5 value signifies a more effective attack.

\subsection{Main Results}
Table \ref{tab:main results of untargeted adversarial attacks} presents the results of the two proposed untargeted adversarial attack methods, with the results without attacks serving as a baseline for comparison. 
Image embedding attacks reduce the average IoU@0.5 value to 23.24\%, while textual bounding box attacks decrease it to an average value of 33.90\%. This difference in effectiveness may be attributed to the fact that image embedding attacks disrupt the original image features, directly impacting the visual grounding capabilities of MLLMs. In contrast, textual bounding box attacks primarily affect the textual generation process of MLLMs, which might not have as significant an effect on tasks that heavily rely on visual input.

Table \ref{tab:main results of targeted adversarial attacks} shows the results of two proposed targeted adversarial attack paradigms. The results without attacks refer to the experiments when original images are used as inputs, with no adversarial perturbations, but with altered labels. Exclusive targeted adversarial attacks can enhance the average IoU@0.5 from 0.13\% to 61.84\%. Meanwhile, Permuted targeted adversarial attacks can improve the IoU@0.5 from 8.89\% to 30.14\%. It can be observed that permuted targeted adversarial attacks are more challenging. The reason is potentially that the area and position of target bounding box area in exclusive targeted adversarial attacks are larger and fixed, whereas the area and position of the target bounding box in permuted targeted adversarial attacks are more refined and random.

\begin{table*}[]
\caption{The IoU@0.5 (\%) of two proposed untargeted adversarial attack methods against MiniGPT-v2 on three datasets. The lower values correspond to a stronger attack. }
\label{tab:main results of untargeted adversarial attacks}
\centering
\small
\setlength\tabcolsep{7.5pt}{
% \resizebox{\linewidth}{!}{
\begin{tabular}{@{}l|cccccccc@{}}
\toprule
\multirow{2}{*}{Method} & \multicolumn{3}{c}{RefCOCO} & \multicolumn{3}{c}{RefCOCO+} & \multicolumn{2}{c}{RefCOCOg}\\ 
 & val & test-A & test-B & val & test-A & test-B & val & test \\
\midrule 
No attack & 84.96 & 89.39 & 82.15 & 76.22 & 82.57 & 70.30 & 81.61 & 82.01 \\
Image embedding attacks & 29.58 & 35.60 & 19.23 & 21.86 & 27.78 & 12.64 & 19.28 & 19.91 \\
Textual bounding box attacks & 43.60 & 49.60 & 36.58 & 36.18 & 42.42 & 28.65 & 36.74 & 37.41 \\
\bottomrule
\end{tabular}}
\end{table*}

\begin{table*}[]
\vspace{-1em}
\caption{The IoU@0.5 (\%) of two proposed targeted adversarial attack paradigms against MiniGPT-v2 on three datasets. The higher values correspond to a stronger attack.}
\label{tab:main results of targeted adversarial attacks}
\centering
\small
\setlength\tabcolsep{9.2pt}{
% \resizebox{\linewidth}{!}{
\begin{tabular}{@{}l|cccccccc@{}}
\toprule
\multirow{2}{*}{Method} & \multicolumn{3}{c}{RefCOCO} & \multicolumn{3}{c}{RefCOCO+} & \multicolumn{2}{c}{RefCOCOg}\\ 
 & val & test-A & test-B & val & test-A & test-B & val & test \\
\midrule 
Exclusive (No attack) & 0.14 & 0.08 & 0.22 & 0.11 & 0.05 & 0.21 & 0.20 & 0.04 \\
Exclusive & 62.12 & 63.94 & 60.98 & 61.93 & 62.90 & 61.11 & 60.96 & 60.77 \\
Permuted (No attack) & 5.69 & 5.17 & 7.43 & 10.65 & 7.87 & 14.1 & 10.09 & 10.15 \\
Permuted & 27.87 & 30.26 & 29.37 & 29.91 & 30.66 & 33.22 & 30.12 & 29.69 \\
\bottomrule
\end{tabular}}
\end{table*}

\section{Related Work}
\subsection{Multimodal Large Language Models}
Multimodal large language models (MLLMs) integrate vision modalities into large language models (LLMs) to extend their capabilities, broadening their scope beyond standard textual understanding and improving their performance across various multimodal tasks \citep{li2022blip,zhu2023minigpt,chen2023minigpt,ma2022visual,ma2022simvtp,ma2023follow}. Recent studies unlock visual grounding capabilities of MLLMs to address localization tasks with region-aware functionalities. Specifically, KOSMOS-2 \citep{peng2023kosmos} and VisionLLM \citep{wang2024visionllm} introduce additional location tokens to the vocabulary, enabling the conversion of coordinates into textual representations, thereby enhancing regional comprehension. Moreover, Shikra \citep{chen2023shikra} and MiniGPT-v2 \citep{chen2023minigpt} directly represent spatial coordinates using natural language, simplifying the integration of spatial data into the model. Despite the effective performance, the security threat for visual grounding of MLLMs, including adversarial learning \citep{goodfellow2014explaining,carlini2019evaluating,dong2023robust}, backdoor learning \citep{li2022backdoor,gao2023backdoor,gao2023imperceptible,bai2023badclip}, poisoning learning \citep{shafahi2018poison}, and Trojan learning \citep{rakin2020tbt,bai2022hardly,bai2023versatile}, has not been studied well.

\subsection{Adversarial Attacks}
Adversarial attacks \citep{goodfellow2014explaining,dong2018boosting,ilyas2018black,zhang2019theoretically,bai2020improving,bai2020targeted,bai2021improving,bai2021targeted} have been widely studied for classification models, where imperceptible and carefully crafted perturbations are applied to input data to mislead the model into producing incorrect predictions. Inspired by the adversarial vulnerability observed in vision tasks, early efforts are devoted to investigating adversarial attacks against   MLLMs \citep{dong2023robust,gao2024inducing,wang2024stop}.  However, the adversarial robustness of MLLMs with visual grounding ability is still under-explored. Since visual grounding reveals the model’s perception process \citep{zhang2018grounding,li2021referring}, it can serve as a good proxy to understand the model behavior before and after the adversarial attacks. To this end, we designing effective attack methods to evaluate the adversarial robustness of MLLMs with grounding ability.

\section{Conclusion}
In this paper, we aim to craft imperceptible perturbations to generate adversarial images, evaluating the adversarial robustness for visual grounding of MLLMs. We propose three adversarial attack paradigms: untargeted adversarial attacks, exclusive targeted adversarial attacks, and permuted targeted adversarial attacks. Comprehensive experimental results on three benchmark datasets, namely RefCOCO, RefCOCO+, and RefCOCOg, demonstrate the effectiveness of our proposed attacks. We hope that our proposed adversarial attacks can serve as a baseline to evaluate the visual grounding ability in adversarial robustness of MLLMs and inspire more research to focus on visual grounding of MLLMs.

\subsection*{ETHICS STATEMENT}
Please note that we restrict all experiments in the laboratory environment and do not support our adversarial attacks in the real scenario.
The purpose of our work is to raise the awareness of the concern in availability of MLLMs and call for  practitioners to pay more attention to the visual grounding in adversarial robustness of MLLMs and model trustworthy deployment.

\subsection*{ACKNOWLEDGEMENT}
This work is supported in part by the National Natural Science Foundation of China under Grant 62171248, Shenzhen Science and Technology Program (JCYJ20220818101012025), and the PCNL KEY project (PCL2023AS6-1).

% in achieving their goals. We hope our proposed adversarial attacks serve as a baseline for evaluating the visual grounding adversarial robustness of MLLMs.

% In this paper, we aim craft an imperceptible perturbation generate adversarial images to evaluate the adversarial robustness for visual grounding of MLLMs. To this end, three adversarial attack paradigms are proposed, including untargeted adversarial attacks, exclusive targeted adversarial attacks, and permuted targeted adversarial attacks. Extensive experimental results demonstrate that our attack can successful achieve their attack goals on three benchmark datasets, \textit{i.e.}, RefCOCO, RefCOCO+, and RefCOCOg. We hope that our proposed adversarial attacks can serve as a baseline to evaluate the visual grounding adversarial robustness of MLLMs.

\clearpage
\bibliography{iclr2024_conference}

\begin{thebibliography}{47}
\providecommand{\natexlab}[1]{#1}
\providecommand{\url}[1]{\texttt{#1}}
\expandafter\ifx\csname urlstyle\endcsname\relax
  \providecommand{\doi}[1]{doi: #1}\else
  \providecommand{\doi}{doi: \begingroup \urlstyle{rm}\Url}\fi

\bibitem[Alayrac et~al.(2022)Alayrac, Donahue, Luc, Miech, Barr, Hasson, Lenc, Mensch, Millican, Reynolds, et~al.]{alayrac2022flamingo}
Jean-Baptiste Alayrac, Jeff Donahue, Pauline Luc, Antoine Miech, Iain Barr, Yana Hasson, Karel Lenc, Arthur Mensch, Katherine Millican, Malcolm Reynolds, et~al.
\newblock Flamingo: a visual language model for few-shot learning.
\newblock In \emph{NeurIPS}, 2022.

\bibitem[Bagdasaryan et~al.(2023)Bagdasaryan, Hsieh, Nassi, and Shmatikov]{bagdasaryan2023ab}
Eugene Bagdasaryan, Tsung-Yin Hsieh, Ben Nassi, and Vitaly Shmatikov.
\newblock (ab) using images and sounds for indirect instruction injection in multi-modal llms.
\newblock \emph{arXiv preprint arXiv:2307.10490}, 2023.

\bibitem[Bai et~al.(2020{\natexlab{a}})Bai, Chen, Li, Wu, Guo, Xia, and Yang]{bai2020targeted}
Jiawang Bai, Bin Chen, Yiming Li, Dongxian Wu, Weiwei Guo, Shu-tao Xia, and En-hui Yang.
\newblock Targeted attack for deep hashing based retrieval.
\newblock In \emph{ECCV}, 2020{\natexlab{a}}.

\bibitem[Bai et~al.(2022{\natexlab{a}})Bai, Gao, Gong, Xia, Li, and Liu]{bai2022hardly}
Jiawang Bai, Kuofeng Gao, Dihong Gong, Shu-Tao Xia, Zhifeng Li, and Wei Liu.
\newblock Hardly perceptible trojan attack against neural networks with bit flips.
\newblock In \emph{ECCV}, 2022{\natexlab{a}}.

\bibitem[Bai et~al.(2022{\natexlab{b}})Bai, Wu, Zhang, Li, Li, and Xia]{bai2021targeted}
Jiawang Bai, Baoyuan Wu, Yong Zhang, Yiming Li, Zhifeng Li, and Shu-Tao Xia.
\newblock Targeted attack against deep neural networks via flipping limited weight bits.
\newblock In \emph{ICLR}, 2022{\natexlab{b}}.

\bibitem[Bai et~al.(2023{\natexlab{a}})Bai, Gao, Min, Xia, Li, and Liu]{bai2023badclip}
Jiawang Bai, Kuofeng Gao, Shaobo Min, Shu-Tao Xia, Zhifeng Li, and Wei Liu.
\newblock Badclip: Trigger-aware prompt learning for backdoor attacks on clip.
\newblock \emph{arXiv preprint arXiv:2311.16194}, 2023{\natexlab{a}}.

\bibitem[Bai et~al.(2023{\natexlab{b}})Bai, Wu, Li, and Xia]{bai2023versatile}
Jiawang Bai, Baoyuan Wu, Zhifeng Li, and Shu-Tao Xia.
\newblock Versatile weight attack via flipping limited bits.
\newblock \emph{IEEE Transactions on Pattern Analysis and Machine Intelligence}, 2023{\natexlab{b}}.

\bibitem[Bai et~al.(2020{\natexlab{b}})Bai, Zeng, Jiang, Wang, Xia, and Guo]{bai2020improving}
Yang Bai, Yuyuan Zeng, Yong Jiang, Yisen Wang, Shu-Tao Xia, and Weiwei Guo.
\newblock Improving query efficiency of black-box adversarial attack.
\newblock In \emph{ECCV}, 2020{\natexlab{b}}.

\bibitem[Bai et~al.(2021)Bai, Zeng, Jiang, Xia, Ma, and Wang]{bai2021improving}
Yang Bai, Yuyuan Zeng, Yong Jiang, Shu-Tao Xia, Xingjun Ma, and Yisen Wang.
\newblock Improving adversarial robustness via channel-wise activation suppressing.
\newblock In \emph{ICLR}, 2021.

\bibitem[Carlini et~al.(2019)Carlini, Athalye, Papernot, Brendel, Rauber, Tsipras, Goodfellow, Madry, and Kurakin]{carlini2019evaluating}
Nicholas Carlini, Anish Athalye, Nicolas Papernot, Wieland Brendel, Jonas Rauber, Dimitris Tsipras, Ian Goodfellow, Aleksander Madry, and Alexey Kurakin.
\newblock On evaluating adversarial robustness.
\newblock \emph{arXiv preprint arXiv:1902.06705}, 2019.

\bibitem[Carlini et~al.(2023)Carlini, Nasr, Choquette-Choo, Jagielski, Gao, Awadalla, Koh, Ippolito, Lee, Tramer, et~al.]{carlini2023aligned}
Nicholas Carlini, Milad Nasr, Christopher~A Choquette-Choo, Matthew Jagielski, Irena Gao, Anas Awadalla, Pang~Wei Koh, Daphne Ippolito, Katherine Lee, Florian Tramer, et~al.
\newblock Are aligned neural networks adversarially aligned?
\newblock \emph{arXiv preprint arXiv:2306.15447}, 2023.

\bibitem[Chen et~al.(2022)Chen, Guo, Yi, Li, and Elhoseiny]{chen2022visualgpt}
Jun Chen, Han Guo, Kai Yi, Boyang Li, and Mohamed Elhoseiny.
\newblock Visualgpt: Data-efficient adaptation of pretrained language models for image captioning.
\newblock In \emph{CVPR}, 2022.

\bibitem[Chen et~al.(2023{\natexlab{a}})Chen, Zhu, Shen, Li, Liu, Zhang, Krishnamoorthi, Chandra, Xiong, and Elhoseiny]{chen2023minigpt}
Jun Chen, Deyao Zhu, Xiaoqian Shen, Xiang Li, Zechun Liu, Pengchuan Zhang, Raghuraman Krishnamoorthi, Vikas Chandra, Yunyang Xiong, and Mohamed Elhoseiny.
\newblock Minigpt-v2: large language model as a unified interface for vision-language multi-task learning.
\newblock \emph{arXiv preprint arXiv:2310.09478}, 2023{\natexlab{a}}.

\bibitem[Chen et~al.(2023{\natexlab{b}})Chen, Zhang, Zeng, Zhang, Zhu, and Zhao]{chen2023shikra}
Keqin Chen, Zhao Zhang, Weili Zeng, Richong Zhang, Feng Zhu, and Rui Zhao.
\newblock Shikra: Unleashing multimodal llm's referential dialogue magic.
\newblock \emph{arXiv preprint arXiv:2306.15195}, 2023{\natexlab{b}}.

\bibitem[Dong et~al.(2018)Dong, Liao, Pang, Su, Zhu, Hu, and Li]{dong2018boosting}
Yinpeng Dong, Fangzhou Liao, Tianyu Pang, Hang Su, Jun Zhu, Xiaolin Hu, and Jianguo Li.
\newblock Boosting adversarial attacks with momentum.
\newblock In \emph{CVPR}, 2018.

\bibitem[Dong et~al.(2023)Dong, Chen, Chen, Fang, Yang, Zhang, Tian, Su, and Zhu]{dong2023robust}
Yinpeng Dong, Huanran Chen, Jiawei Chen, Zhengwei Fang, Xiao Yang, Yichi Zhang, Yu~Tian, Hang Su, and Jun Zhu.
\newblock How robust is google's bard to adversarial image attacks?
\newblock \emph{arXiv preprint arXiv:2309.11751}, 2023.

\bibitem[Gao et~al.(2023{\natexlab{a}})Gao, Bai, Wu, Ya, and Xia]{gao2023imperceptible}
Kuofeng Gao, Jiawang Bai, Baoyuan Wu, Mengxi Ya, and Shu-Tao Xia.
\newblock Imperceptible and robust backdoor attack in 3d point cloud.
\newblock \emph{IEEE Transactions on Information Forensics and Security}, 19:\penalty0 1267--1282, 2023{\natexlab{a}}.

\bibitem[Gao et~al.(2023{\natexlab{b}})Gao, Bai, Gu, Yang, and Xia]{gao2023backdoor}
Kuofeng Gao, Yang Bai, Jindong Gu, Yong Yang, and Shu-Tao Xia.
\newblock Backdoor defense via adaptively splitting poisoned dataset.
\newblock In \emph{CVPR}, 2023{\natexlab{b}}.

\bibitem[Gao et~al.(2024{\natexlab{a}})Gao, Bai, Gu, Xia, Torr, Li, and Liu]{gao2024inducing}
Kuofeng Gao, Yang Bai, Jindong Gu, Shu-Tao Xia, Philip Torr, Zhifeng Li, and Wei Liu.
\newblock Inducing high energy-latency of large vision-language models with verbose images.
\newblock In \emph{ICLR}, 2024{\natexlab{a}}.

\bibitem[Gao et~al.(2024{\natexlab{b}})Gao, Gu, Bai, Xia, Torr, Liu, and Li]{gao2024energy}
Kuofeng Gao, Jindong Gu, Yang Bai, Shu-Tao Xia, Philip Torr, Wei Liu, and Zhifeng Li.
\newblock Energy-latency manipulation of multi-modal large language models via verbose samples.
\newblock \emph{arXiv preprint arXiv:2404.16557}, 2024{\natexlab{b}}.

\bibitem[Goodfellow et~al.(2015)Goodfellow, Shlens, and Szegedy]{goodfellow2014explaining}
Ian~J Goodfellow, Jonathon Shlens, and Christian Szegedy.
\newblock Explaining and harnessing adversarial examples.
\newblock In \emph{ICLR}, 2015.

\bibitem[Ilyas et~al.(2018)Ilyas, Engstrom, Athalye, and Lin]{ilyas2018black}
Andrew Ilyas, Logan Engstrom, Anish Athalye, and Jessy Lin.
\newblock Black-box adversarial attacks with limited queries and information.
\newblock In \emph{ICML}, 2018.

\bibitem[Kazemzadeh et~al.(2014)Kazemzadeh, Ordonez, Matten, and Berg]{kazemzadeh2014referitgame}
Sahar Kazemzadeh, Vicente Ordonez, Mark Matten, and Tamara Berg.
\newblock Referitgame: Referring to objects in photographs of natural scenes.
\newblock In \emph{EMNLP}, 2014.

\bibitem[Li et~al.(2021)Li, Selvaraju, Gotmare, Joty, Xiong, and Hoi]{li2021align}
Junnan Li, Ramprasaath Selvaraju, Akhilesh Gotmare, Shafiq Joty, Caiming Xiong, and Steven Chu~Hong Hoi.
\newblock Align before fuse: Vision and language representation learning with momentum distillation.
\newblock In \emph{NeurIPS}, 2021.

\bibitem[Li et~al.(2022{\natexlab{a}})Li, Li, Xiong, and Hoi]{li2022blip}
Junnan Li, Dongxu Li, Caiming Xiong, and Steven Hoi.
\newblock Blip: Bootstrapping language-image pre-training for unified vision-language understanding and generation.
\newblock In \emph{ICML}, 2022{\natexlab{a}}.

\bibitem[Li et~al.(2023)Li, Li, Savarese, and Hoi]{li2023blip}
Junnan Li, Dongxu Li, Silvio Savarese, and Steven Hoi.
\newblock Blip-2: Bootstrapping language-image pre-training with frozen image encoders and large language models.
\newblock In \emph{ICML}, 2023.

\bibitem[Li \& Sigal(2021)Li and Sigal]{li2021referring}
Muchen Li and Leonid Sigal.
\newblock Referring transformer: A one-step approach to multi-task visual grounding.
\newblock In \emph{NeurIPS}, 2021.

\bibitem[Li et~al.(2022{\natexlab{b}})Li, Jiang, Li, and Xia]{li2022backdoor}
Yiming Li, Yong Jiang, Zhifeng Li, and Shu-Tao Xia.
\newblock Backdoor learning: A survey.
\newblock \emph{IEEE Transactions on Neural Networks and Learning Systems}, 2022{\natexlab{b}}.

\bibitem[Liu et~al.(2023)Liu, Li, Wu, and Lee]{liu2023visual}
Haotian Liu, Chunyuan Li, Qingyang Wu, and Yong~Jae Lee.
\newblock Visual instruction tuning.
\newblock \emph{arXiv preprint arXiv:2304.08485}, 2023.

\bibitem[Ma et~al.(2022{\natexlab{a}})Ma, Wang, Wu, Lyu, Chen, Li, and Qiao]{ma2022visual}
Yue Ma, Yali Wang, Yue Wu, Ziyu Lyu, Siran Chen, Xiu Li, and Yu~Qiao.
\newblock Visual knowledge graph for human action reasoning in videos.
\newblock In \emph{ACM MM}, 2022{\natexlab{a}}.

\bibitem[Ma et~al.(2022{\natexlab{b}})Ma, Yang, Shan, and Li]{ma2022simvtp}
Yue Ma, Tianyu Yang, Yin Shan, and Xiu Li.
\newblock Simvtp: Simple video text pre-training with masked autoencoders.
\newblock \emph{arXiv preprint arXiv:2212.03490}, 2022{\natexlab{b}}.

\bibitem[Ma et~al.(2024)Ma, He, Cun, Wang, Shan, Li, and Chen]{ma2023follow}
Yue Ma, Yingqing He, Xiaodong Cun, Xintao Wang, Ying Shan, Xiu Li, and Qifeng Chen.
\newblock Follow your pose: Pose-guided text-to-video generation using pose-free videos.
\newblock In \emph{AAAI}, 2024.

\bibitem[Madry et~al.(2018)Madry, Makelov, Schmidt, Tsipras, and Vladu]{madry2017towards}
Aleksander Madry, Aleksandar Makelov, Ludwig Schmidt, Dimitris Tsipras, and Adrian Vladu.
\newblock Towards deep learning models resistant to adversarial attacks.
\newblock In \emph{ICLR}, 2018.

\bibitem[Mao et~al.(2016)Mao, Huang, Toshev, Camburu, Yuille, and Murphy]{mao2016generation}
Junhua Mao, Jonathan Huang, Alexander Toshev, Oana Camburu, Alan~L Yuille, and Kevin Murphy.
\newblock Generation and comprehension of unambiguous object descriptions.
\newblock In \emph{CVPR}, 2016.

\bibitem[OpenAI(2023)]{openai2023gpt4}
OpenAI.
\newblock Gpt-4 technical report.
\newblock 2023.

\bibitem[Peng et~al.(2023)Peng, Wang, Dong, Hao, Huang, Ma, and Wei]{peng2023kosmos}
Zhiliang Peng, Wenhui Wang, Li~Dong, Yaru Hao, Shaohan Huang, Shuming Ma, and Furu Wei.
\newblock Kosmos-2: Grounding multimodal large language models to the world.
\newblock \emph{arXiv preprint arXiv:2306.14824}, 2023.

\bibitem[Qi et~al.(2023)Qi, Huang, Panda, Wang, and Mittal]{qi2023visual}
Xiangyu Qi, Kaixuan Huang, Ashwinee Panda, Mengdi Wang, and Prateek Mittal.
\newblock Visual adversarial examples jailbreak large language models.
\newblock \emph{arXiv preprint arXiv:2306.13213}, 2023.

\bibitem[Rakin et~al.(2020)Rakin, He, and Fan]{rakin2020tbt}
Adnan~Siraj Rakin, Zhezhi He, and Deliang Fan.
\newblock Tbt: Targeted neural network attack with bit trojan.
\newblock In \emph{CVPR}, 2020.

\bibitem[Shafahi et~al.(2018)Shafahi, Huang, Najibi, Suciu, Studer, Dumitras, and Goldstein]{shafahi2018poison}
Ali Shafahi, W~Ronny Huang, Mahyar Najibi, Octavian Suciu, Christoph Studer, Tudor Dumitras, and Tom Goldstein.
\newblock Poison frogs! targeted clean-label poisoning attacks on neural networks.
\newblock In \emph{NeurIPS}, 2018.

\bibitem[Wang et~al.(2024{\natexlab{a}})Wang, Chen, Chen, Wu, Zhu, Zeng, Luo, Lu, Zhou, Qiao, et~al.]{wang2024visionllm}
Wenhai Wang, Zhe Chen, Xiaokang Chen, Jiannan Wu, Xizhou Zhu, Gang Zeng, Ping Luo, Tong Lu, Jie Zhou, Yu~Qiao, et~al.
\newblock Visionllm: Large language model is also an open-ended decoder for vision-centric tasks.
\newblock In \emph{NeurIPS}, 2024{\natexlab{a}}.

\bibitem[Wang et~al.(2024{\natexlab{b}})Wang, Han, Chen, Xue, Ding, Xiao, Tresp, Torr, and Gu]{wang2024stop}
Zefeng Wang, Zhen Han, Shuo Chen, Fan Xue, Zifeng Ding, Xun Xiao, Volker Tresp, Philip Torr, and Jindong Gu.
\newblock Stop reasoning! when multimodal llms with chain-of-thought reasoning meets adversarial images.
\newblock \emph{arXiv preprint arXiv:2402.14899}, 2024{\natexlab{b}}.

\bibitem[Yang et~al.(2024)Yang, Bai, Jia, Liu, Cao, and Yu]{yang2024cheating}
Dingcheng Yang, Yang Bai, Xiaojun Jia, Yang Liu, Xiaochun Cao, and Wenjian Yu.
\newblock Cheating suffix: Targeted attack to text-to-image diffusion models with multi-modal priors.
\newblock \emph{arXiv preprint arXiv:2402.01369}, 2024.

\bibitem[Yu et~al.(2016)Yu, Poirson, Yang, Berg, and Berg]{yu2016modeling}
Licheng Yu, Patrick Poirson, Shan Yang, Alexander~C Berg, and Tamara~L Berg.
\newblock Modeling context in referring expressions.
\newblock In \emph{ECCV}, 2016.

\bibitem[Zhang et~al.(2018)Zhang, Niu, and Chang]{zhang2018grounding}
Hanwang Zhang, Yulei Niu, and Shih-Fu Chang.
\newblock Grounding referring expressions in images by variational context.
\newblock In \emph{CVPR}, 2018.

\bibitem[Zhang et~al.(2019)Zhang, Yu, Jiao, Xing, El~Ghaoui, and Jordan]{zhang2019theoretically}
Hongyang Zhang, Yaodong Yu, Jiantao Jiao, Eric Xing, Laurent El~Ghaoui, and Michael Jordan.
\newblock Theoretically principled trade-off between robustness and accuracy.
\newblock In \emph{ICML}, 2019.

\bibitem[Zhao et~al.(2023)Zhao, Pang, Du, Yang, Li, Cheung, and Lin]{zhao2023evaluating}
Yunqing Zhao, Tianyu Pang, Chao Du, Xiao Yang, Chongxuan Li, Ngai-Man Cheung, and Min Lin.
\newblock On evaluating adversarial robustness of large vision-language models.
\newblock \emph{arXiv preprint arXiv:2305.16934}, 2023.

\bibitem[Zhu et~al.(2023)Zhu, Chen, Shen, Li, and Elhoseiny]{zhu2023minigpt}
Deyao Zhu, Jun Chen, Xiaoqian Shen, Xiang Li, and Mohamed Elhoseiny.
\newblock Minigpt-4: Enhancing vision-language understanding with advanced large language models.
\newblock 2023.

\end{thebibliography}
\bibliographystyle{iclr2024_conference}

\end{document}